\documentclass{article}
\usepackage{spconf,amsmath,graphicx,hyperref}
\usepackage{siunitx}
\usepackage{cite}
\usepackage{amsmath,amssymb,amsfonts}
\usepackage{algorithmic}
\usepackage{textcomp}
\usepackage{xcolor}
\usepackage{multirow}

\usepackage[T1]{fontenc}
\usepackage{multirow} 
\usepackage{listings}
\usepackage{pifont}
\usepackage{graphicx,verbatim}
\usepackage{booktabs}
\usepackage{makecell}
\usepackage{lipsum} 
\usepackage{subcaption}
\usepackage{caption}
\usepackage{float}
\usepackage{pifont}
\usepackage{booktabs}
\usepackage{marvosym}

\usepackage{orcidlink}
\usepackage[marginal]{footmisc}
\title{VasGuideNet: Vascular Topology-Guided Couinaud Liver Segmentation with Structural Contrastive Loss}
\name{
Chaojie Shen$^{1\dag}$, 
Jingjun Gu$^{3\dag}$, 
Zihao Zhao$^{2\dag}$, 
Ruocheng Li$^{2}$\orcidlink{0009-0000-4669-8385}, 
Cunyuan Yang$^{2}$, 
Jiajun Bu$^{2(}$\href{mailto:bjj@zju.edu.cn}{\textsuperscript{\Letter}}$^)$,
Lei Wu$^{2(}$\href{mailto:shenhai1895@zju.edu.cn}{\textsuperscript{\Letter}}$^)$\orcidlink{0000-0002-0980-3016}
}
\address{
         $^{1}$ Pujian Medical Technology Co., Ltd., Hangzhou, China \\
         $^{2}$ College of Computer Science and Technology, Zhejiang University, Hangzhou, China\\
         $^{3}$ Keyi College of Zhejiang Sci-tech University, ShaoXing, China,  \\
         qacket@126.com, \{gjj, 22221130, lirc, cunyuany, bjj, shenhai1895\}@zju.edu.cn,
         }

\begin{document}
%\ninept
%
\maketitle
\def\thefootnote{$\dag$}\footnotetext{Equal contribution; \Letter~corresponding author.
}

\begin{abstract}
Accurate Couinaud liver segmentation is critical for preoperative surgical planning and tumor localization.
However, existing methods primarily rely on image intensity and spatial location cues, without explicitly modeling vascular topology. As a result, they often produce indistinct boundaries near vessels and show limited generalization under anatomical variability.
We propose \textbf{VasGuideNet}, the first Couinaud segmentation framework explicitly guided by vascular topology. Specifically, skeletonized vessels, Euclidean distance transform (EDT)--derived geometry, and k-nearest neighbor (kNN) connectivity are encoded into topology features using Graph Convolutional Networks (GCNs). These features are then injected into a 3D encoder--decoder backbone via a cross-attention fusion module. To further improve inter-class separability and anatomical consistency, we introduce a \textbf{Structural Contrastive Loss (SCL)} with a global memory bank.
On Task08\_HepaticVessel and our private LASSD dataset, VasGuideNet achieves Dice scores of 83.68\% and 76.65\% with RVDs of 1.68 and 7.08, respectively. It consistently outperforms representative baselines including UNETR, Swin UNETR, and G-UNETR++, delivering higher Dice/mIoU and lower RVD across datasets, demonstrating its effectiveness for anatomically consistent segmentation. Code is available at \url{https://github.com/Qacket/VasGuideNet.git}.
\end{abstract}

\begin{keywords}
Couinaud segmentation, vascular topology, graph convolutional networks (GCNs), structural contrastive loss
\end{keywords}

\begin{figure*}[htb]
  \centering
  \includegraphics[width=\linewidth]{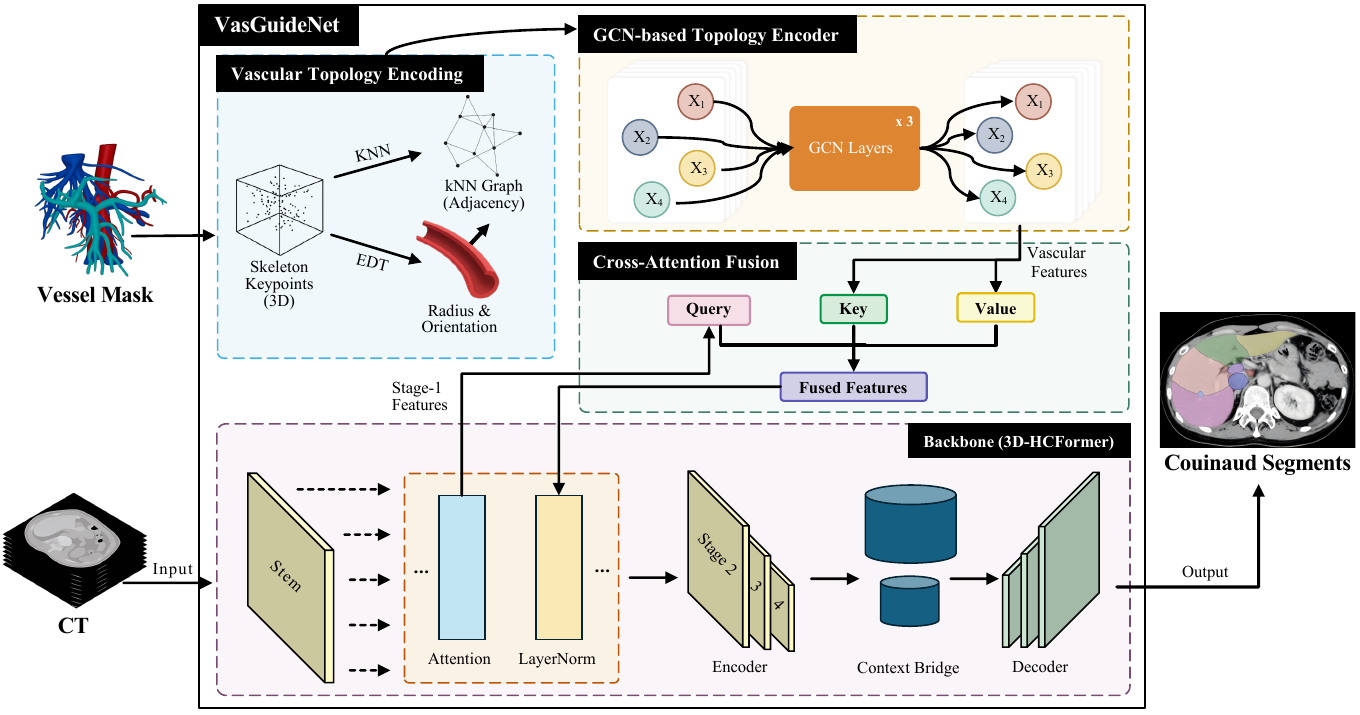}
    \caption{
        Overview of VasGuideNet for Couinaud liver segmentation. 
    }
  \label{fig:framework}
\end{figure*}

\section{Introduction}
Couinaud segmentation divides the liver into eight functionally independent segments based on the distribution of the portal and hepatic veins~\cite{couinaud1957liver,healey1953anatomy}.
It provides a critical anatomical basis for surgical planning and postoperative evaluation. 
Accurate Couinaud segmentation on preoperative CT is essential for guiding hepatectomy, especially in cases with anatomical variation, as it enables surgeons to determine safe resection boundaries while preserving sufficient functional parenchyma.

Automatic Couinaud segmentation remains highly challenging. 
Adjacent segments often show similar image intensity and texture patterns, while boundaries near vascular structures are weak or ambiguous. 
Inter-patient variability, contrast phase differences, and domain shifts across centers further exacerbate these difficulties. 
Traditional atlas-based and registration methods~\cite{linguraru2009hepatic} attempt to impose prior anatomical constraints. However, they struggle with large variations and require extensive preprocessing, which limits clinical applicability.

Recent deep learning methods have significantly advanced liver and Couinaud segmentation. 
Architectures such as 3D U-Net~\cite{cciccek20163d}, nnU-Net~\cite{isensee2021nnunet}, and Swin UNETR~\cite{hatamizadeh2022swin} achieve strong single-center results but lose boundary precision near vessels.
Transformer-based approaches such as LiverFormer~\cite{qiu2023liverformer} and 3D-HCFormer improve long-range dependency modeling, yet still lack explicit vascular topology encoding. 
% Consequently, these methods often suffer from inter-segment confusion in anatomically complex cases and show limited robustness across acquisition settings and anatomical variability.

From the perspective of Couinaud’s anatomical theory, liver segmentation is essentially a functional partitioning process governed by vascular distribution. 
% Experienced clinicians identify segments by observing the spatial arrangement of hepatic vessels, suggesting that vascular topology is a natural and powerful prior for anatomically consistent segmentation.
Previous vascular-related works, such as vesselness filters~\cite{frangi1998multiscale}, topology-preserving segmentation~\cite{wang2020topology}, and vascular graph representations~\cite{chen2021vesselgraph}, show the potential of vascular priors, yet they are not explicitly incorporated into Couinaud frameworks.

Moreover, most existing losses such as Dice or Focal primarily optimize voxel-wise accuracy and address class imbalance. 
However, they do not leverage anatomical priors. 
This limitation is particularly problematic where inter-segment boundaries are ambiguous near vessels. 
Contrastive learning~\cite{he2020momentum} has recently been adapted to medical imaging~\cite{chaitanya2020contrastive,hu2022structural} to enforce feature separability and exploit structure, indicating that structural priors can enhance generalization with limited annotations. 
These insights motivate the design of a Structural Contrastive Loss (SCL) that leverages vascular guidance for improved robustness under anatomical and acquisition variability.

We study Couinaud segmentation with expert-annotated vessel masks available at both training and inference, focusing on vascular guidance rather than vessel prediction.
Based on this setting, we present \textbf{VasGuideNet}, a vascular topology--guided Couinaud segmentation framework.
The method explicitly encodes vascular skeletons, Euclidean distance transforms, and vessel connectivity graphs. These features are integrated into the segmentation backbone via a GCN-based cross-attention fusion module. Moreover, a Structural Contrastive Loss with a global memory bank is proposed to enforce compact intra-class clusters and well-separated inter-class boundaries.
The main contributions of this work are threefold:
\begin{itemize}
    \item We introduce explicit vascular topology modeling into Couinaud segmentation, enabling boundary refinement near vascular structures.
    \item We design a GCN-based cross-attention fusion module that integrates topology-aware features with image semantics for anatomically consistent predictions.
    \item We propose a Structural Contrastive Loss (SCL) with a memory bank to enhance inter-class separability and improve robustness under anatomical and acquisition variability.
\end{itemize}

\section{Method}

\subsection{Network Overview}
% As shown in Fig.~\ref{fig:framework}, our \textbf{VasGuideNet} integrates vascular topology into Couinaud liver segmentation through four modules: 
% (i) Vascular Topology Encoding, 
% (ii) GCN-based Topology Encoder, 
% (iii) Cross-Attention Fusion, and 
% (iv) a Backbone (3D-HCFormer).
As shown in Fig.~\ref{fig:framework}, \textbf{VasGuideNet} injects vascular topology features into a 3D-HCFormer backbone via cross-attention.
The backbone consists of a Stem for downsampling and channel projection, four encoder stages, a Context Bridge for long-range dependencies, and a Decoder. 
% Features from the topology branch are injected into the backbone via cross-attention, guiding image representations with vascular priors to generate the final Couinaud segmentation map.
We assume an expert-annotated vessel mask is available at both training and inference; no vessel prediction is used at test time.

\subsection{Vascular Topology Encoding}
Given a vessel segmentation mask, we first perform 3D skeletonization to obtain the vessel centerline and uniformly sample $N$ skeleton keypoints $\{p_i \in \mathbb{R}^3\}_{i=1}^N$. 
We then compute the Euclidean distance transform (EDT) of the binary vessel mask. 
At each keypoint $p_i$, we extract the local radius $r_i \approx D(p_i)$ and a normalized orientation vector $n_i = \nabla D(p_i)/\|\nabla D(p_i)\|$. 
Finally, we construct a $k$-nearest neighbor (kNN) graph over the keypoints to obtain the adjacency matrix $A$. 
The initial node feature for each keypoint is computed as
\begin{equation}
h_i^{(0)} = \mathrm{MLP}([p_i, r_i, n_i]),
\end{equation}
yielding a feature matrix
\begin{equation}
X = [h_i^{(0)}]_{i=1}^N \in \mathbb{R}^{N \times d_0}.
\end{equation}

\subsection{GCN-based Topology Encoder}
To capture the connectivity and branching patterns of the hepatic vessel tree, we apply a stack of $L$ graph convolutional layers on the input graph $(X, A)$:
\begin{equation}
H^{(\ell+1)} = \sigma(\hat{A} H^{(\ell)} W^{(\ell)}), \quad H^{(0)} = X,
\end{equation}
where $\hat{A}$ is the normalized adjacency matrix with self-loops, $W^{(\ell)}$ are learnable weights, and $\sigma(\cdot)$ denotes a nonlinear activation. The output node embeddings
\begin{equation}
Z = H^{(L)} \in \mathbb{R}^{N \times d}
\end{equation}
are defined as vascular topology features.

\subsection{Vascular Topology Injection}
The topology features $Z$ are injected into the backbone via a Transformer-style cross-attention mechanism. Specifically, the Stage-1 encoder output from the backbone is used as Query tokens $Q = F W_Q$, while the topology features provide Key and Value tokens $K = Z W_K$ and $V = Z W_V$. Cross-attention is computed as
\begin{equation}
\mathrm{Attn}(Q, K, V) = \mathrm{softmax}\left(\frac{QK^\top}{\sqrt{d_k}}\right)V,
\end{equation}
producing fused features that are propagated through the encoder, Context Bridge, and decoder to generate the final prediction. 
% This design guides image features toward vessel-informed boundaries, improving delineation near hepatic veins and portal branches.

\subsubsection{Ablation fusion variants}
\label{sssec:fusion}
For ablation studies, we evaluate three alternative fusion strategies: 
\textbf{No Injection}, which removes the topology branch entirely; 
\textbf{Channel Concatenation}, which concatenates topology and image features before the decoder; and 
\textbf{Distance Bias}, which uses the EDT map as an additive bias to decoder attention weights.

\subsection{Loss Function}

\begin{figure}[htb]
  \centering
  \includegraphics[width=\linewidth]{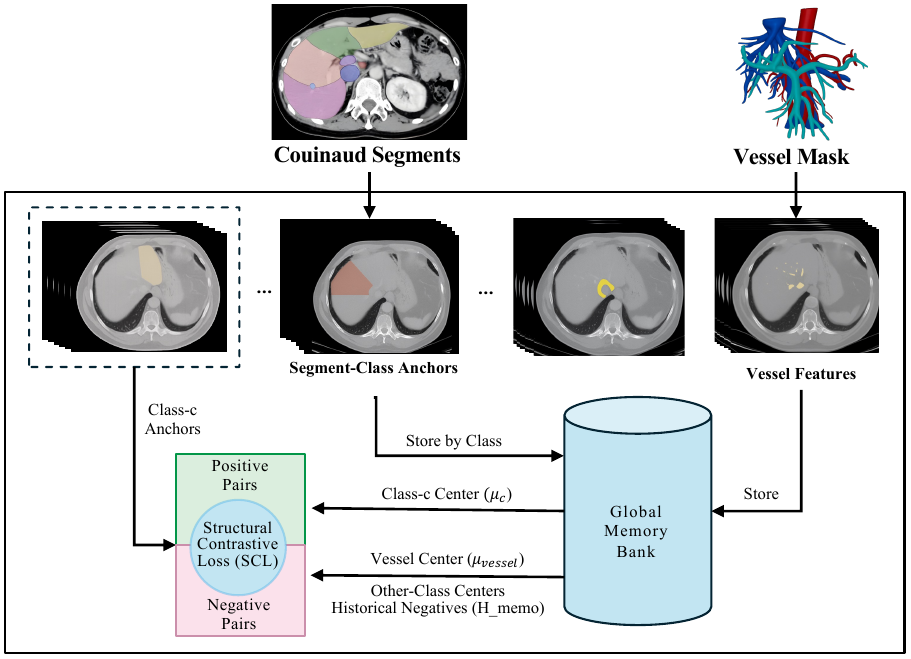}
    \caption{
        Structural Contrastive Loss (SCL). 
    }
  \label{fig:scl}
\end{figure}

% Standard losses such as Dice or Focal primarily optimize voxel-wise accuracy and address class imbalance.
% However, they do not explicitly encode anatomical structure.
To enhance inter-class separability and anatomical consistency, we introduce a \textbf{Structural Contrastive Loss (SCL)} guided by vascular topology, as illustrated in Fig.~\ref{fig:scl}. SCL selects high-confidence voxels as anchors. 
These anchors are contrasted against feature centers of other segments, the vessel center, and historical negatives stored in a global memory bank.

\subsubsection{Anchor Feature Selection}
For each voxel $i$, let $\mathrm{Conf}_i$ denote the predicted confidence. We select anchors above the $95$th percentile $Q^{\mathrm{Conf}}_{95}$:
\begin{equation}
A_c = \{ F_i \mid \mathrm{Conf}_i > Q^{\mathrm{Conf}}_{95}, \ M_i = c \},
\end{equation}
where $F_i$ is the feature at voxel $i$, $M_i$ is its ground-truth label, and $A_c$ denotes the anchor set for class $c$.

\subsubsection{Negative Sample Construction}
For each class $c$, the negative set is defined as
\begin{equation}
B_c = \{ \mu_k \mid k \neq c \} \cup \mu_{\text{vessel}} \cup \{ H_{\text{memo},k} \mid k \neq c \},
\end{equation}
where $\mu_k$ is the feature center of class $k$, $\mu_{\text{vessel}}$ is the vessel center, and $H_{\text{memo},k}$ denotes historical negatives for class $k$.

\subsubsection{Structural Contrastive Loss}
The SCL objective is formulated as
\begin{equation}
\mathcal{L}_{\text{SCL}} = \frac{1}{N} \sum_{c=1}^{N} \frac{1}{|A_c|} \sum_{a_i \in A_c} 
-\log \frac{\exp(\mathrm{sim}(a_i, \mu_c)/\tau)}{\sum_{n \in B_c} \exp(\mathrm{sim}(a_i, n)/\tau)},
\end{equation}
where $\mathrm{sim}(\cdot, \cdot)$ denotes cosine similarity and $\tau$ is a temperature parameter. This objective encourages compact intra-class clusters and well-separated inter-class boundaries.

\subsubsection{Memory Bank Update Strategy}
\label{sssec:strategy}
We maintain a global memory bank for negative samples and investigate two update strategies: 
\textbf{FIFO}, a first-in-first-out queue that ensures temporal diversity but may discard high-quality entries; and 
\textbf{CATS (Confidence-Aware Time-Slice)}, which preferentially retains high-confidence items and replaces low-confidence ones, improving stability and contrast effectiveness.

\section{Experiments}
We conducted experiments on two liver CT datasets. 
% The first is the \textbf{Task08\_HepaticVessel} dataset~\cite{antonelli2022medical} from the MICCAI 2020 Abdominal Challenge, which comprises 443 CT scans and is publicly available at \url{http://medicaldecathlon.com/}.
The first is the \textbf{Task08\_HepaticVessel} dataset~\cite{antonelli2022medical} (443 CT scans).
Couinaud labels follow prior annotations~\cite{tian2019automatic}. 
The second dataset is the private \textbf{LASSD} collection, consisting of 147 CT scans with expert-annotated Couinaud labels provided by our collaborating hospital. 
For Task08\_HepaticVessel, we use the official vessel annotations for both training and testing; for LASSD, vessel masks are manually annotated by experienced clinical experts and used in the same manner during training and inference.

Segmentation performance is reported in terms of Dice Similarity Coefficient (DSC), mean Intersection over Union (mIoU), and Relative Volume Difference (RVD), which together assess overlap accuracy, boundary precision, and volume consistency.

\begin{table}[htbp]
\caption{Comparison with state-of-the-art methods (5-fold CV).}
\centering
\resizebox{\linewidth}{!}{%
\begin{tabular}{l S[table-format=2.2(2)] S[table-format=2.2(2)] S[table-format=2.2(2)]}
\toprule
\textbf{Method} & {\textbf{DSC (\%)}$\uparrow$} & {\textbf{mIoU (\%)}$\uparrow$} & {\textbf{RVD (\%)}$\downarrow$} \\
\midrule
\multicolumn{4}{c}{\textbf{Task08\_HepaticVessel}} \\
\midrule
G-UNETR++   & 59.32 \pm 2.26 & 42.78 \pm 2.29 & 34.11 \pm 3.89 \\
SegMamba    & 79.05 \pm 3.34 & 67.61 \pm 3.66 & 29.34 \pm 15.34 \\
Unetr       & 81.31 \pm 1.60 & 57.76 \pm 27.19 & 27.50 \pm 37.38 \\
LiverFormer & 71.51 \pm 1.87 & 41.60 \pm 19.10 & 49.24 \pm 31.83 \\
SegFormer   & 79.51 \pm 1.76 & 66.79 \pm 2.08  &  4.80 \pm 0.85 \\
3D-HCFormer & 82.23 \pm 0.87 & 70.61 \pm 1.11  &  3.50 \pm 1.56 \\
\textbf{VasGuideNet} & \bfseries 83.68 \pm 0.82 & \bfseries 72.67 \pm 1.06 & \bfseries 1.68 \pm 0.29 \\
\midrule
\multicolumn{4}{c}{\textbf{LASSD}} \\
\midrule
G-UNETR++   & 65.14 \pm 1.55 & 50.15 \pm 1.49 & 30.49 \pm 8.51 \\
SegMamba    & 69.00 \pm 6.28 & 55.66 \pm 6.71 & 46.39 \pm 27.09 \\
Unetr       & 70.03 \pm 6.11 & 56.48 \pm 6.34 & 33.04 \pm 23.21 \\
LiverFormer & 57.23 \pm 4.51 & 41.40 \pm 3.86 & 63.58 \pm 12.30 \\
SegFormer   & 73.09 \pm 3.53 & 59.78 \pm 2.87 & 11.44 \pm 2.12 \\
3D-HCFormer & 75.00 \pm 4.01 & 62.31 \pm 3.50 & 11.28 \pm 4.79 \\
\textbf{VasGuideNet} & \bfseries 76.65 \pm 3.33 & \bfseries 64.40 \pm 2.89 & \bfseries 7.08 \pm 1.55 \\
\bottomrule
\end{tabular}%
}
\label{tab:sota}
\end{table}

\subsection{Comparison with State-of-the-Art}
We compare \textbf{VasGuideNet} with representative 3D segmentation baselines, including UNETR~\cite{hatamizadeh2022unetr}, SegFormer~\cite{xie2021segformer}, SegMamba~\cite{xing2024segmamba}, G-UNETR++~\cite{lee2025g}, LiverFormer~\cite{qiu2023liverformer}, and 3D-HCFormer. Table~\ref{tab:sota} summarizes the results. 

On \textbf{Task08\_HepaticVessel}, VasGuideNet achieves the best Dice (83.68\%), mIoU (72.67\%), and the lowest RVD (1.68). 
Compared with 3D-HCFormer, it improves Dice by +1.45, mIoU by +2.06, and reduces RVD by 1.82, indicating more accurate boundaries and better volume calibration.
On the private \textbf{LASSD} dataset, it also outperforms 3D-HCFormer (76.65\% vs.\ 75.00\% Dice, 64.40\% vs.\ 62.31\% mIoU, 7.08 vs.\ 11.28 RVD). 
These results highlight the advantage of explicit vascular-topology modeling under anatomical variability.

\subsection{Ablation Studies}

\subsubsection{Vascular Topology Injection Strategies}
We compare the proposed GCN--cross-attention fusion against three alternatives: \emph{None}, \emph{Channel Concatenation}, and \emph{Distance Bias} (Table~\ref{tab:fusion}; definitions in Section~\ref{sssec:fusion}). Relative to \emph{None}, VasGuideNet improves Dice by +1.45, mIoU by +2.06, and reduces RVD by 1.82. Channel Concatenation provides moderate gains but still lags behind VasGuideNet, while Distance Bias slightly lowers RVD with limited improvement in Dice/mIoU. These results confirm that explicit topology modeling via GCN--cross-attention is crucial for anatomical consistency.

\begin{table}[!t]
\caption{Comparison of Multi-Source Fusion Strategies.}
\centering
\resizebox{\linewidth}{!}{%
\begin{tabular}{l 
                S[table-format=2.2(2)]
                S[table-format=2.2(2)]
                S[table-format=2.2(2)]}
\toprule
\textbf{Fusion Strategy} & {\textbf{DSC (\%)}$\uparrow$} & {\textbf{mIoU (\%)}$\uparrow$} & {\textbf{RVD (\%)}$\downarrow$} \\
\midrule
None                  & 82.23 \pm 0.87 & 70.61 \pm 1.11  &  3.50 \pm 1.56 \\
Channel Concatenation & 83.23 \pm 1.05 & 71.91 \pm 1.25 & 2.23 \pm 0.44 \\
Distance Bias         & 82.25 \pm 0.63 & 70.59 \pm 0.72 & 2.91 \pm 1.05 \\
\textbf{VasGuideNet} & \bfseries 83.68 \pm 0.82 & \bfseries 72.67 \pm 1.06 & \bfseries 1.68 \pm 0.29 \\
\bottomrule
\end{tabular}%
}
\label{tab:fusion}
\end{table}

\subsubsection{Structural Contrastive Loss (SCL)}

We compare a baseline without SCL, a \emph{FIFO} strategy, and the proposed \emph{CATS} update strategy (Table~\ref{tab:scl}; see Section~\ref{sssec:strategy}).
Adding SCL markedly improves performance over \emph{None} by +2.79 Dice, +4.00 mIoU, and reduces RVD by 1.14. The confidence-aware \emph{CATS} further surpasses \emph{FIFO} with +0.59 Dice, +0.91 mIoU, and $-$0.30 RVD, confirming that prioritizing high-confidence features yields more compact clusters and more accurate volumes.

\begin{table}[!t]
\caption{SCL update strategies.}
\centering
\resizebox{\linewidth}{!}{%
\begin{tabular}{l
                S[table-format=2.2(2)]
                S[table-format=2.2(2)]
                S[table-format=2.2(2)]}
\toprule
\textbf{Update Strategy} & {\textbf{DSC (\%)}$\uparrow$} & {\textbf{mIoU (\%)}$\uparrow$} & {\textbf{RVD (\%)}$\downarrow$} \\
\midrule
None & 80.30 \pm 0.92 & 67.76 \pm 1.06 & 3.12 \pm 0.56 \\
FIFO & 83.09 \pm 0.72 & 71.76 \pm 0.86 & 1.98 \pm 0.15 \\
\textbf{CATS} & \bfseries 83.68 \pm 0.82 & \bfseries 72.67 \pm 1.06 & \bfseries 1.68 \pm 0.29 \\
\bottomrule
\end{tabular}%
}
\label{tab:scl}
\end{table}

\section{Conclusion}

We present \textbf{VasGuideNet}, which encodes skeleton/EDT/kNN vessel topology via GCN and injects it into 3D-HCFormer by cross-attention with SCL.
Experiments on Task08\_HepaticVessel and the private LASSD dataset demonstrate consistent improvements in Dice and mIoU, as well as reduced RVD, compared with representative baselines. These results highlight the potential of vascular topology guidance for anatomically consistent segmentation, with direct applicability to preoperative surgical planning.

\section{Acknowledgments}
This work was supported by Pujian Medical Technology Co., Ltd., Hangzhou, China.

\section{Compliance with Ethical Standards}
The study on Task08\_HepaticVessel was conducted retrospectively using human subject data made available in open access by Medical Segmentation Decathlon. Ethical approval was not required as confirmed by the license attached with the open access data.
The study on LASSD was performed in line with the principles of the Declaration of Helsinki. Approval was granted by the Ethics Committee of Zhejiang University (No. 202201059).

\bibliographystyle{IEEEbib}
\bibliography{strings,refs}
\end{document}